\documentclass[runningheads]{llncs}
\usepackage{graphicx}
\usepackage{amsmath}
\usepackage{multirow}
\usepackage{amssymb}
\usepackage{algorithm}
\usepackage{algorithmic}
\usepackage{color}
\usepackage{makecell}
\usepackage{pifont}
\usepackage[misc,geometry]{ifsym}

\begin{document}
%

\titlerunning{LGPMA}
%
\title{LGPMA: Complicated Table Structure Recognition with Local and Global Pyramid Mask Alignment\thanks{Both \emph{L. Qiao} and \emph{Z. Li} contributed equally.\emph{ Z. Cheng} is the corresponding author. Supported by National Key R\&D Program of China(Grant No. 2018YFC0831601).}}

\author{Liang Qiao\inst{1}
        \and Zaisheng Li\inst{1}
         \and Zhanzhan Cheng\inst{2,1}(\Letter)
         \and Peng Zhang\inst{1}
         \and Shiliang Pu\inst{1}
         \and Yi Niu\inst{1}
         \and Wenqi Ren\inst{1}
         \and Wenming Tan\inst{1}
         \and Fei Wu\inst{2}}
\authorrunning{L. Qiao, Z. Li, Z. Cheng et al.}
%
\institute{Hikvision Research Institute, China \\
\email{\{qiaoliang6, lizaisheng, chengzhanzhan, zhangpeng23, pushiliang.hri, niuyi, renwenqi, tanwenming\}@hikvision.com}
\and
Zhejiang University, China \\
\email{wufei@cs.zju.edu.cn}}

\maketitle              

\begin{abstract}
Table structure recognition is a challenging task due to the various structures and complicated cell spanning relations.
Previous methods handled the problem starting from elements in different granularities (rows/columns, text regions), which somehow fell into the issues like lossy heuristic rules or neglect of empty cell division.
Based on table structure characteristics, we find that obtaining the aligned bounding boxes of text region can effectively maintain the entire relevant range of different cells.
However, the aligned bounding boxes are hard to be accurately predicted due to the visual ambiguities.
In this paper, we aim to obtain more reliable aligned bounding boxes by fully utilizing the visual information from both text regions in proposed local features and cell relations in global features. Specifically, we propose the framework of \emph{Local and Global Pyramid Mask Alignment}, which adopts the soft pyramid mask learning mechanism in both the local and global feature maps. It allows the predicted boundaries of bounding boxes to break through the limitation of original proposals. A pyramid mask re-scoring module is then integrated to compromise the local and global information and refine the predicted boundaries.
Finally, we propose a robust table structure recovery pipeline to obtain the final structure, in which we also effectively solve the problems of empty cells locating and division. Experimental results show that the proposed method achieves competitive and even new state-of-the-art performance on several public benchmarks. The code is available in \url{https://github.com/hikopensource/DAVAR-Lab-OCR/tree/main/demo/table\_recognition/lgpma}.
\keywords{Table Structure Recognition \and Aligned Bounding Box  \and Empty Cell.}
\end{abstract}

\section{Introduction}
Table is one of the rich-information data formats in many real documents like financial statements, scientific literature, purchasing lists, etc. Besides the text content, the table structure is vital for people to do the key information extraction. Thus, table structure recognition~\cite{kieninger1998table,nishida2017understanding,WangPH04,schreiber2017deepdesrt,gobel2013icdar,gao2019icdar,zhong2019image} becomes one of the important techniques in current document understanding systems.

From the global perspective, early table structure recognition processes usually depend on the detection of the grid's boundaries~\cite{liu2008identifying,liu2009improving}. However, these methods can not handle tables without grid boundaries, such as three-line tables.
Though recent works~\cite{schreiber2017deepdesrt,paliwal2019tablenet,siddiqui2019rethinking,siddiqui2019deeptabstr}  attempt to predict row/column regions or even invisible grid lines~\cite{tensmeyer2019deep}, they are limited to handle tables that cross span multiple rows/columns.
The row/column splitting operation might also cut cells that contain text in multiple lines.

\begin{figure*}[t]
\begin{center}
\includegraphics[width=1\textwidth]{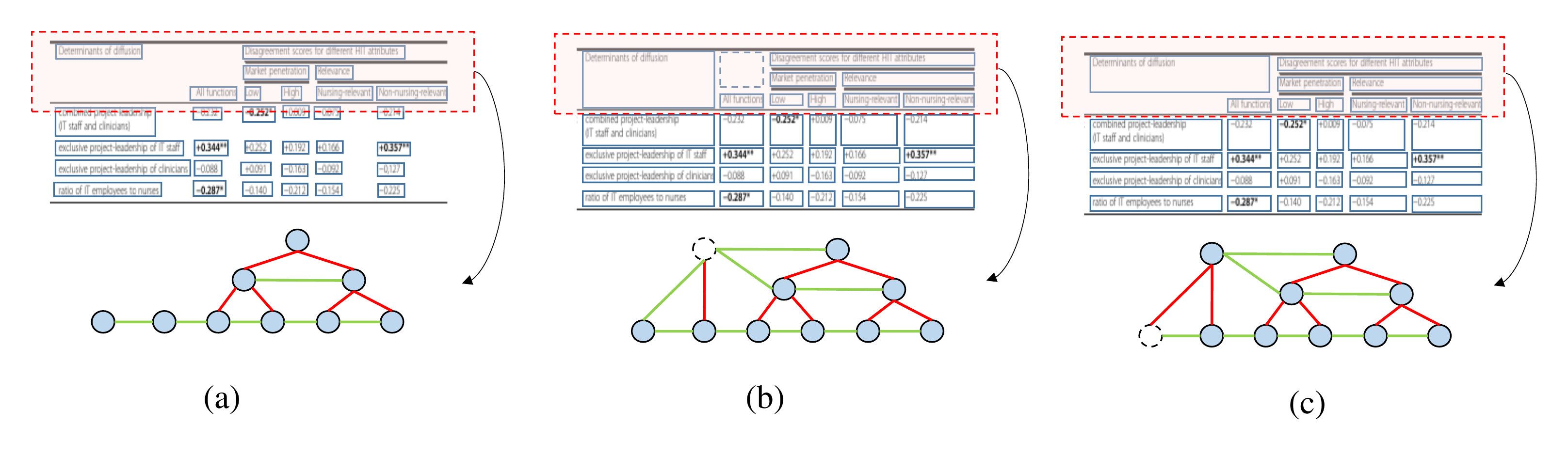}\\
\end{center}
\caption{ (a) The visualized results without considering empty cells. (b) The ground-truth of aligned bounding boxes and nodes relations. (c) A false example due to the ambiguity between the empty cell and cross-column cell.
The cells and their relations are represented as nodes and connected lines (red: vertical, green: horizontal). Empty cells are displayed in dashed circles.}
\label{intro}
\end{figure*}

Another group of methods solves the above problems in a bottom-up way to firstly detect the text blocks' positions and then recover the bounding-boxes' relations by heuristic rules~\cite{zheng2020global} or GNN(Graph Neural Networks)~\cite{ScarselliGTHM09,li2020gfte,chi2019complicated,qasim2019rethinking,raja2020table}.
However, rules designed based on bounding boxes of text regions are vulnerable to handling complicated matching situations. GNN-based methods not only bring extra network cost but also depend on more expensive training cost such as the data volume. Another issue is that these methods are difficult to obtain the empty cells because they usually fall into the visual ambiguity problem with the cross-row/column cells. The prediction of empty cells directly affects the correctness of table structure, as illustrated in Figure \ref{intro}(a).
Moreover, how to split or merge these empty regions is still a challenging problem that cannot be neglected, because different division results will generate different editable areas when the image is transferred into digital format.

Notice that the structure of the table itself is a human-made rule-based data form. Under the situation that tables are without visual rotation or perspective transformation, if we could obtain all of the perfect aligned cell regions rather than the text regions~\cite{raja2020table}, the structure inference will be easy and almost lossless, as illustrated in Figure \ref{intro}(b). Nevertheless, acquiring such information is not easy. On the one hand, the annotations of text regions~\cite{gobel2013icdar,chi2019complicated,zhong2019image} are much easier to get than cell regions. On the other hand, the aligned boxes are difficult to be accurately learned since there is usually no visible texture of boundaries at the region's periphery.  Multi-row/column cells are easy to be confused with the empty cell regions. For example, in Figure \ref{intro}(c), the network usually falls into the situation that the predicted aligned boxes are not large enough and results in the wrong cell matching. Although~\cite{raja2020table} designs an alignment loss to assist the bounding boxes learning, it only considers the relative relations between boxes and fails to capture the cell's absolute coverage area.

In this paper, we aim to train the network to obtain more reliable aligned cell regions and solve the problems of empty cell generation and partition in one model.
Observing that people perceive visual information from both local text region and global layout when they read, we propose a uniform table structure recognition framework to compromise the benefits from both local and global information, called LGPMA (Local and Global Pyramid Mask Alignment) Network. Specifically, the model simultaneously learns a local Mask-RCNN-based~\cite{he2017mask} aligned bounding boxes detection task and a global segmentation task. In both tasks, we adopt the pyramid soft mask supervision~\cite{liu2019pyramid} to help obtain more accurate aligned bounding boxes. In LGPMA, the local branch (LPMA) acquires more reliable text region information through visible texture perceptron, while the global branch (GPMA) can learn more legible spatial information of cells' range or division.
The two branches help the network learn better-fused features via jointly learning and effectively refine the detected aligned bounding boxes through a proposed mask re-scoring strategy. Based on the refined results, we design a robust and straightforward table structure recovery pipeline, which can effectively locate empty cells and precisely merge them according to the guidance of global segmentation.

The major contributions of this paper are as follows:
(1) We propose a novel framework called LGPMA Network that compromises the visual features from both local and global perspectives. The model makes full use of the information from the local and global features through a proposed mask re-scoring strategy, which can obtain more reliable aligned cell regions.
(2) We introduce a uniform table structure recovering pipeline, including cell matching, empty cell searching, and empty cell merging. Both non-empty cells and empty cells can be located and split efficaciously.
(3) Extensive experiments show that our method achieves competitive and even state-of-the-art results on several popular benchmarks.

\section{Related Works}
Traditional table recognition researches mainly worked with hand-crafted features and heuristic rules~\cite{Itonori93,kieninger1998table,WangPH04,liu2008identifying,liu2009improving,DoushP10}. These methods are mostly applied to simple table structures or specific data formats, such as PDFs. The early techniques about table detection and recognition can be found in the comprehensive survey~\cite{ZanibbiBC04}.
With the great success of deep neural network in computer vision field, works began to focus on the image-based table with more general structures~\cite{nishida2017understanding,schreiber2017deepdesrt,qasim2019rethinking,khan2019table,LiCHWZL20,xue2019res2tim,prasad2020cascadetabnet,li2020gfte,tensmeyer2019deep}. According to the basic components granularities, we roughly divide previous methods into two types: global-object-based methods and local-object-based methods.

\textbf{Global-object-based methods} mainly focus on the characteristics of global table components and mostly started from row/column or grid boundaries detection. Works of~\cite{schreiber2017deepdesrt,siddiqui2019deeptabstr,siddiqui2019rethinking} firstly obtain the rows and columns regions using the detection or segmentation models and then intersect these two regions to obtain the grids of cells. \cite{paliwal2019tablenet} handles the table detection and table recognition tasks in an end-to-end manner by the table region mask learning and table's row/column mask learning. \cite{tensmeyer2019deep} detects the rows and columns by learning the interval areas' segmentation between rows/columns and then predicting the indicator to merge the separated cells.

There also exist some methods~\cite{LiCHWZL20,zhong2019image} that directly perceive the whole image information and output table structures as text sequence in an encoder-decoder framework. Although these methods look graceful and entirely avoid human being involved, the models are usually challenging to be trained and rely on a large amount of training data. Global-object-based methods usually have difficulties in handling various complicated table structures, such as cells spanning multiple rows/columns or containing text in multi-lines.

\textbf{Local-object-based methods} begin from the smallest fundamental element, cells. Given the cell-level text region annotation, the text detection task is relatively easy to finish by the general detection methods like Yolo~\cite{Redmon2016You}, Faster R-CNN~\cite{2015Faster}, etc. After that, a group of methods~\cite{xue2019res2tim,prasad2020cascadetabnet,zheng2020global} tries to recover the cell relations based on some heuristic rules and algorithms. Another type of methods~\cite{KociTL018,chi2019complicated,li2020gfte,qasim2019rethinking,raja2020table} treat the detected boxes as nodes in a graph and attempt to predict the relations based on techniques of Graph Neural Networks~\cite{ScarselliGTHM09}. \cite{li2020gfte} predicts the relations between nodes in three classes (the horizontal connection, the vertical connection, no connection) using several features such as visual features, text positions, word embedding, etc.  \cite{chi2019complicated} adopts graph attention mechanism to enhance the predicting accuracy. \cite{qasim2019rethinking} alleviates the problem of large graph nodes numbers by the pair sampling strategy. The above three works~\cite{li2020gfte,chi2019complicated,qasim2019rethinking} also published new table datasets for this research area. Since there is no empty cell detected, local-object based-methods usually fall into empty cell ambiguity.

In this paper, we try to compromise the advantages of both global and local features.   Based on the local detection results, we integrate the global information to refine the detected bounding boxes and provide a straightforward guide for empty cell division.

\section{Methodology}
\begin{figure*}[t]
\begin{center}
\includegraphics[width=1.0\textwidth,height=5.5cm]{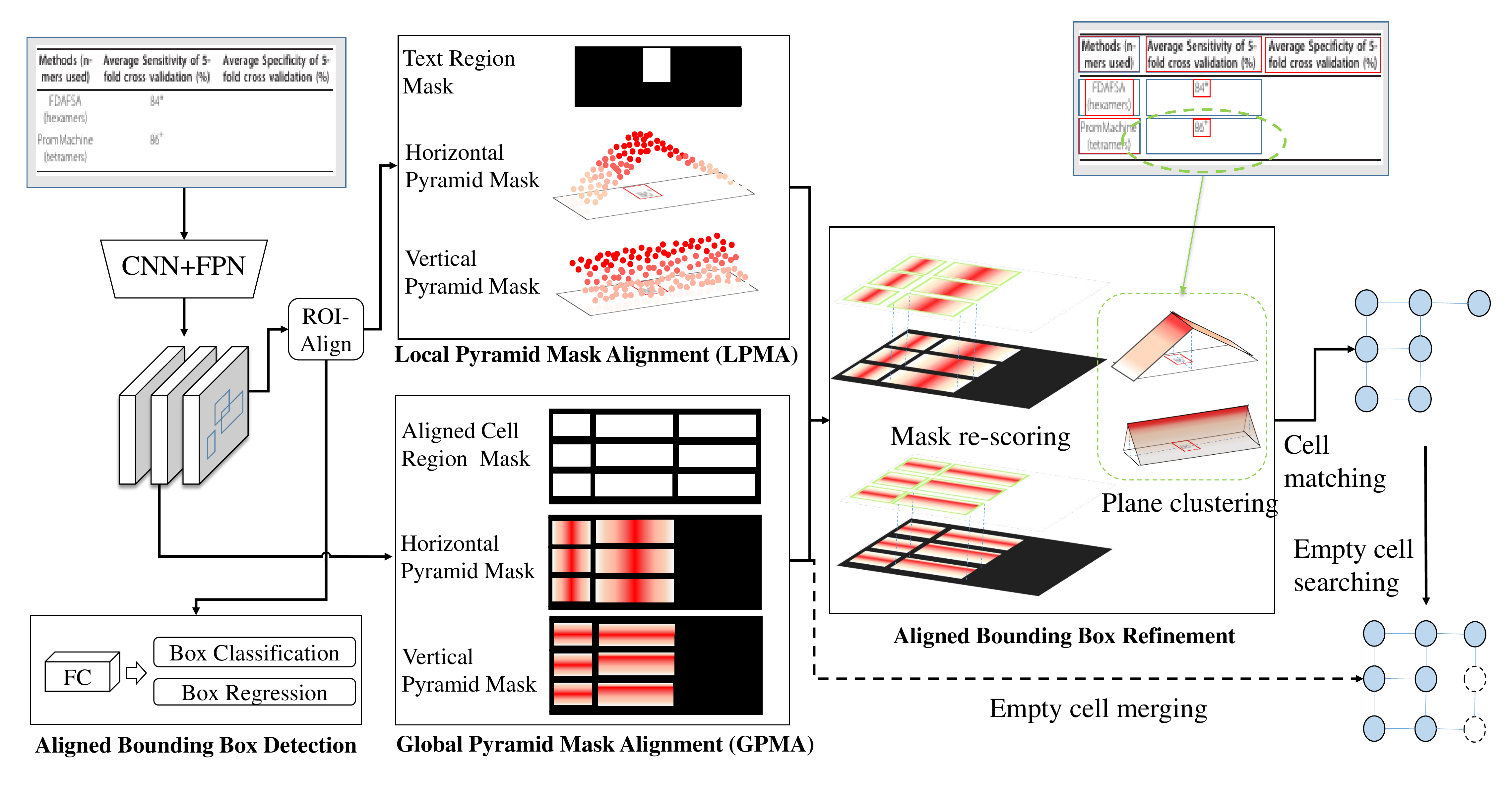}\\
\end{center}
\caption{The workflow of LGPMA. The network simultaneously learns a local aligned bounding boxes detection task (LPMA) and a global segmentation task (GPMA). We adopt the pyramid mask learning mechanisms in both branches and use a mask re-scoring strategy to refine the predicted bounding boxes. Finally, the table structure can be uniformly recovered by a pipeline, including cell matching, empty cell searching, and empty cell merging.  }
\label{framework}
\end{figure*}

\subsection{Overview}
We propose the model LGPMA, whose overall workflow is shown in Figure \ref{framework}.

The model is built based on the existing Mask-RCNN~\cite{he2017mask}. The bounding box branch directly learns the detection task of aligned bounding boxes for non-empty cells. The network simultaneously learns a Local Pyramid Mask Alignment (LPMA) task based on the local feature extracted by the RoI-Align operation and a Global Pyramid Mask Alignment (GPMA) task based on the global feature map.

In LPMA, in addition to the binary segmentation task that learns the text region mask, the network is also trained with the pyramid soft mask supervision in both horizontal and vertical directions.

In GPMA, the network learns a global pyramid mask for all aligned bounding boxes of non-empty cells. To obtain more information about empty cell splitting, the network also learns the global binary segmentation task that considers both non-empty and empty cells.

A pyramid mask re-scoring module is then adopted to refine the predicted pyramid labels. The accurate aligned bounding boxes can be obtained by the process of plane clustering. Finally, a uniform structure recovering pipeline containing cell matching, empty cell searching, empty cell merging is integrated to obtain the final table structure.

\subsection{Aligned Bounding Box Detection}

The difficulty of accurate text region matching mainly comes from the covered range gap between text regions and the real cell regions. Real cell regions may contain empty spaces for row/column alignment, especially for those cells crossing span multiple rows/columns.
Inspired by~\cite{raja2020table,xue2019res2tim}, with the annotations of text regions and row/column indices, we can easily generate the aligned bounding box annotations according to the maximum box height/width in each row/column. The regions of aligned bounding boxes approximately equal to that of real cells.
For the table images in print format and without visual rotation or perspective transformation, if we could obtain the aligned cell regions and assume there is no empty cell, it is easy to infer the cell relations according to the coordinate overlapping information in horizontal and vertical directions.

We adopt Mask-RCNN~\cite{he2017mask} as the base model. In the bounding box branch, the network is trained based on the aligned bounding box supervision.
However, the aligned bounding box learning is not easy because cells are easy to be confused with empty regions. Motivated by the advanced pyramid mask text detector~\cite{liu2019pyramid}, we find that using the soft-label segmentation may break through the proposed bounding box's limitation and provide more accurate aligned bounding boxes.
To fully utilize the visual features from both local texture and global layout, we propose to learn the pyramid mask alignment information in these two folds simultaneously.

\subsection{Local Pyramid Mask Alignment}
In the mask branch, the model is trained to learn both a binary segmentation task and a pyramid mask regression task, which we call Local Pyramid Mask Alignment (LPMA).

The binary segmentation task is the same as the original model, in which only the text region is labeled as 1 and others are labeled as 0. The detected mask regions can be used in the following text recognition task.
\begin{figure*}[t]
\begin{center}
\includegraphics[width=0.8\textwidth, height=1.7cm]{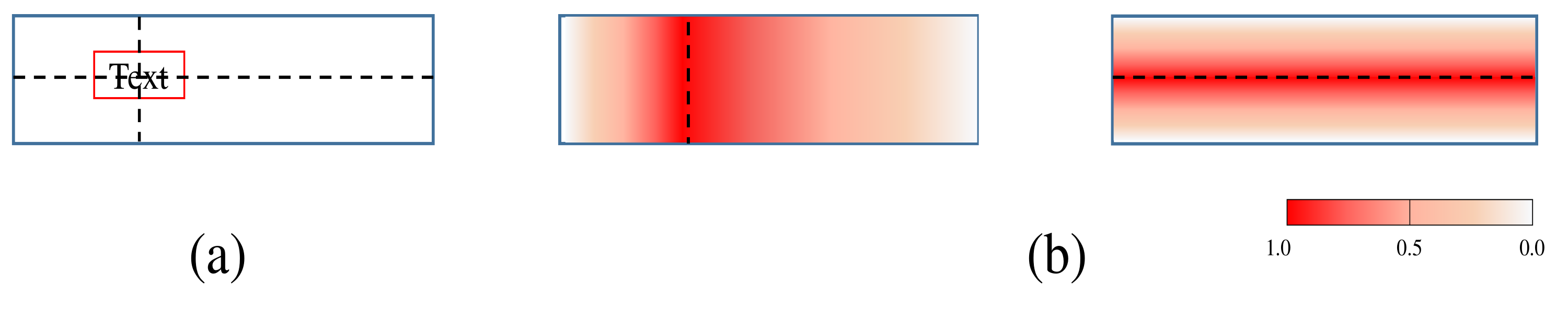}\\
\end{center}
\caption{(a) shows the original aligned bounding box (blue) and text region box (red). (b) shows the pyramid mask labels in horizontal and vertical direction, respectively.}
\label{fig:2}
\end{figure*}

For the pyramid mask regression, we assign the pixels in the proposal bounding box regions with the soft-label in both horizontal and vertical directions, as shown in Figure \ref{fig:2}. The middle point of text will have the largest regressed target 1. Specifically, we assume the proposed aligned bounding box has the shape of $H\times W$. The top-left point and bottom right point of the text region are denoted as $\{(x_1, y_1),(x_2, y_2)\}$, respectively, where  $0$$\le$$x_1$$<$$x_2$$\le W$ and  $0$$\le$$y_1$$<$$y_2$$\le H$. Therefore, the target of the pyramid mask is in shape $\mathbb{R}^{2 \times H \times W}$$\in$$[0,1]$, in which the two channels represent the target map of the horizontal mask and vertical mask, respectively. For every pixel $(h,w)$, these two targets can be formed as:

\begin{equation}
t_h^{(w,h)}=\left\{
\begin{array}{lr}
\ w/x_{mid}    \quad     w \le x_{mid} \\
\frac{W-w}{W-x_{mid}} \quad   w > x_{mid}  \\
\end{array}
\right. , \quad
t_v^{(w,h)} = \left\{
\begin{array}{lr}
\  h/y_{mid}   \quad   h \le y_{mid} \\
\frac{H-h}{H-y_{mid}} \quad   h > y_{mid} \\
\end{array}
\right.,
\end{equation}
where $0$$\le$$w$$<$$W$, $0$$\le$$h$$<$$H$, and $x_{mid}$$=$$\frac{x_1+x_2}{2}$, $y_{mid}=\frac{y_1+y_2}{2}$. In this way, every pixel in the proposal region takes part in predicting the boundaries.

\subsection{Global Pyramid Mask Alignment}
Although LPMA allows the predicted mask to break through the proposal bounding boxes, the local region's receptive fields are limited. To determine the accurate coverage area of a cell, the global feature might also provide some visual clues. Inspired by~\cite{zhou2017east,qiao2020text}, learning the offsets of each pixel from a global view could help locate more accurate boundaries. However, bounding boxes in cell-level might be varied in width-height ratios, which leads to the unbalance problem in regression learning. Therefore, we use the pyramid labels as the regressing targets for each pixel, named Global Pyramid Mask Alignment (GPMA).

Like LPMA, the GPMA learns two tasks simultaneously: a global segmentation task and a global pyramid mask regression task. In the global segmentation task, we directly segment all aligned cells, including non-empty and empty cells. The ground-truth of empty cells are generated according to the maximum height/width of the non-empty cells in the same row/column. Notice that only this task learns empty cell division information since empty cells don't have visible text texture that might influence the region proposal networks to some extent.
We want the model to capture the most reasonable cell division pattern during the global boundary segmentation according to the human's reading habit, which is reflected by the manually labeled annotations.
For the global pyramid mask regression, since only the text region could provide the information of distinct `mountain top,' all non-empty cells will be assigned with the soft labels similar to LPMA.
All of the ground-truths of aligned bounding boxes in GPMA will be shrunk by 5\% to prevent boxes from overlapping.

\subsection{Optimization}
The proposed network is trained end-to-end with multiple optimization tasks. The global optimization can be written as:
\begin{equation}
\mathcal{L} = \mathcal{L}_{rpn} + \lambda_{1}(\mathcal{L}_{cls}+\mathcal{L}_{box}) + \lambda_{2}(\mathcal{L}_{mask}+\mathcal{L}_{LPMA}) + \lambda_{3}(\mathcal{L}_{seg}+\mathcal{L}_{GPMA}),
\end{equation}
where $\mathcal{L}_{rpn}, \mathcal{L}_{cls}, \mathcal{L}_{box}, \mathcal{L}_{mask}$ are the same losses with that of Mask-RCNN, which represent the region proposal network loss, the bounding box classification loss, the bounding boxes regression loss and the segmentation loss of mask in proposals, respectively. $\mathcal{L}_{seg}$ is the global binary segmentation loss that is implemented in Dice coefficient loss~\cite{milletari2016v}, $\mathcal{L}_{LPMA}$ and $\mathcal{L}_{GPMA}$ are the pyramid label regression losses which are optimized by pixel-wise L1 loss. $\lambda_{1},\lambda_{2},\lambda_{3}$ are weighted parameters.

\subsection{Inference}
The inference process can be described in two stages. We first obtain the refined aligned bounding boxes according to the pyramid mask prediction and then generate the final table structure by the proposed structure recovery pipeline.

\subsubsection{Aligned Bounding Box Refine.}
In addition to the benefits generated via joint training, the local and global features also exhibit various advantages in object perceiving ~\cite{xie2018scene}.  In our setting, we find that local features predict more reliable text region masks, while global prediction can provide more credible long-distance visual information. To compromise both levels' merits, we propose a pyramid mask re-scoring strategy to compromise predictions from LPMA and GPMA. For any proposal region with local pyramid mask prediction, we add the information that comes from the global pyramid mask to adjust these scores. We use some dynamic weights to balance the impacts from LPMA and GPMA.

Specifically, for a predicted aligned bounding box $B$$=$$\{(x_{1},y_{1}),(x_2, y_2)\}$, we firstly obtain the bounding box of the text region mask, denoted as $B_t$$=$$\{(x'_{1},y'_{1})$, $(x'_2, y'_2)\}$. Then, we can find a matched connected region $P$$=$$\{p_1,$$p_2$$,$$...,$$p_n\}$ in the global segmentation map, where $p$$=$$(x, y)$ represents a pixel. We use $P_o= \{ p| x_1 \le p.x \le x_2, y_1 \le p.y \le y_2, \forall p \in P \}$ to represent the overlap region. Then the predicted pyramid label of point $(x,y) \in P_o $ can be re-scored as follows.
\begin{equation}
    F(x) = \left\{
 \begin{array}{lr}
    \frac{x-x_{1}}{x_{mid}-x_1} F_{hor}^{(L)}(x,y) + \frac{x_{mid}-x}{x_{mid}-x_1} F_{hor}^{(G)}(x,y)  \qquad x_1 \le x \le x_{mid} \\
    \frac{x-x_{2}}{x_{mid}-x_2} F_{hor}^{(L)}(x,y) + \frac{x_{mid}-x}{x_{mid}-x_2} F_{hor}^{(G)}(x,y)   \qquad x_{mid} < x \le x_2 \\
 \end{array}
 \right. ,
\end{equation}
\begin{equation}
    F(y) = \left\{
 \begin{array}{lr}
    \frac{y-y_{1}}{y_{mid}-y_1} F_{ver}^{(L)}(x,y) + \frac{y_{mid}-y}{y_{mid}-y_1} F_{ver}^{(G)}(x,y)  \qquad y_1 \le y \le y_{mid} \\
    \frac{y-y_{2}}{y_{mid}-y_2} F_{ver}^{(L)}(x,y) + \frac{y_{mid}-y}{y_{mid}-y_2} F_{ver}^{(G)}(x,y)   \qquad y_{mid} < y \le y_2 \\
 \end{array}
 \right. ,
\end{equation}
where $x_{mid}=\frac{x'_1+x'_2}{2}, y_{mid}=\frac{y'_1+y'_2}{2}, F_{hor}^{(L)}(x,y), F_{hor}^{(G)}(x,y), F_{ver}^{(L)}(x,y), F_{ver}^{(G)}(x,y)$ are the local horizontal, global horizontal, local vertical and global vertical pyramid label prediction, respectively.

Next, for any proposal region, the horizontal and vertical pyramid mask labels (corresponding to the $z$-coordinate) can be used to fit two planes in the 3-dimensional space, respectively. All the four planes' intersection lines with the zero plane are the refined boundaries.
For example, to refine the right boundary of the aligned box, we select all pixels that $P_r=\{p| x_{mid} \le p.x \le x_2, p\in P_o\}$ with the refined pyramid mask prediction $F(x,y)$ to fit the plane. If we formed the plane as $ax+by+c-z=0$, using the least square method, the problem is equal to minimize the equation of:

\begin{equation}
\min \sum_{y_i=y_1}^{y_2} \sum_{x_i=x_{mid}}^{x_2} (ax_i+by_i+c-F(x_i,y_i))^2, \qquad \forall p=(x_i,y_i) \in P_r.
\end{equation}
The parameters of $a,b,c$ can be calculated by the matrix as follows:
\begin{equation}
\begin{split}
\left(\begin{matrix} a \\ b \\ c \end{matrix}\right) = \begin{bmatrix} \sum x_i^2 & \ \sum x_iy_i &  \sum x_i \\ \ \ \sum x_iy_i & \sum y_i^2 & \sum y_i \\
\sum x_i & \sum y_i & ||P_o|| \\
 \end{bmatrix} ^{-1} \left(\begin{matrix} x_iF(x_i,y_i)\\ y_iF(x_i,y_i) \\
F(x_i,y_i) \\
 \end{matrix} \right),
\end{split}
\end{equation}
where $||.||$ is the set size.
Then we calculate the intersection line between the fitting plane with the plane of $z=0$. Given that the bounding boxes are axis-aligned, we calculate the refined x-coordinate as the average value:
\begin{equation}
x_{refine} = -\frac{1}{y_2-y_1+1} \sum_{y_i=y_1}^{y_2} \frac{by_i+c}{a}
\end{equation}

Similarly, we can obtain the other three refined boundaries. Notice that the refining process can optionally be conducted iteratively refer to~\cite{liu2019pyramid}.
\subsubsection{Table Structure Recovery.}

\begin{figure*}[t]
\begin{center}
\includegraphics[width=0.9\textwidth, height=4.3cm]{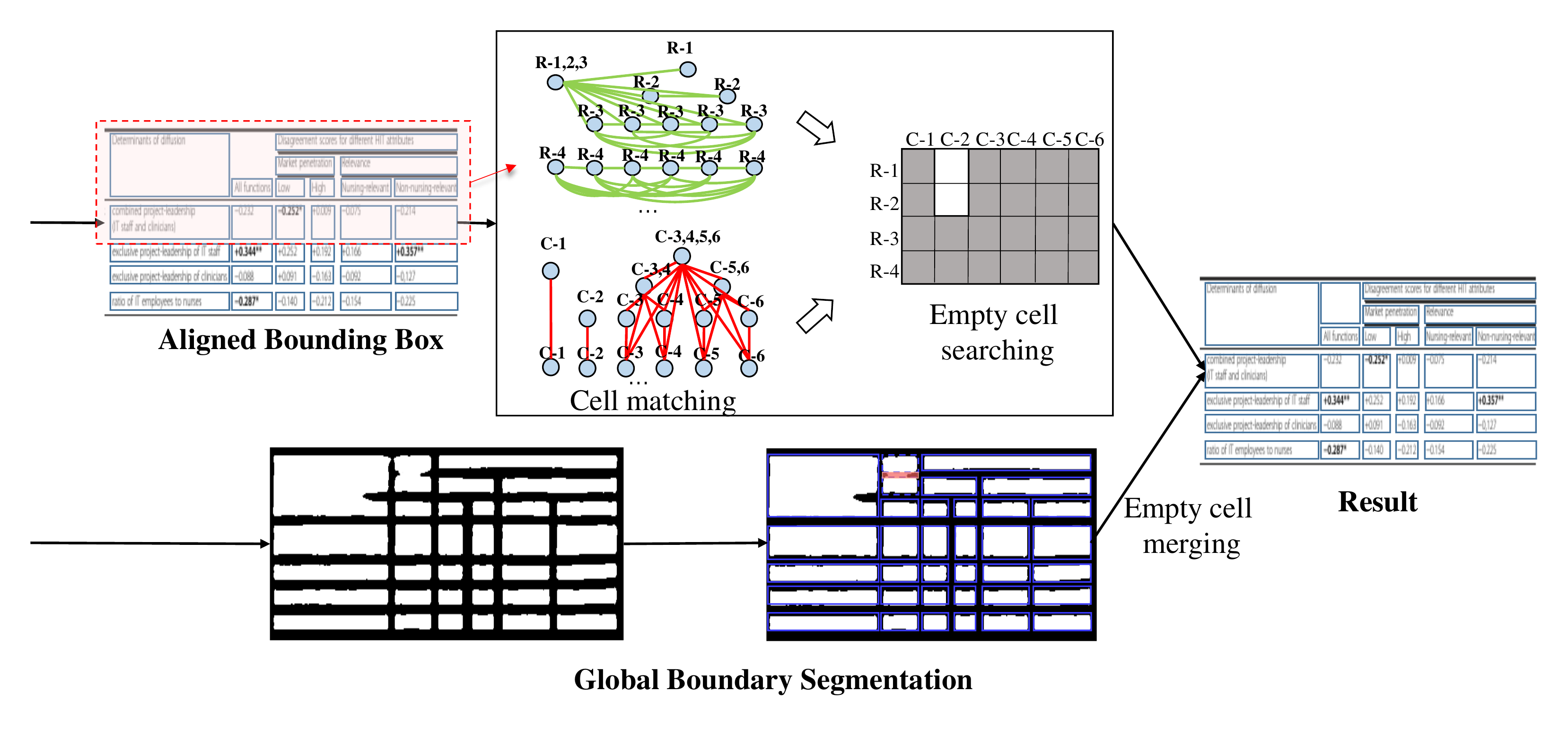}\\
\end{center}
\caption{The illustration of table structure recovery pipeline.}
\label{fig:4}
\end{figure*}

Based on the refined aligned bounding boxes, the table structure recovery pipeline aims to obtain the final table structure, including three steps: \emph{cell matching, empty cell searching and empty cell merging}, as illustrated in Figure \ref{fig:4}.

\emph{Cell Matching.}
In the situation that all of the aligned bounding boxes are axis-aligned, the cells matching process is pretty simple but robust. Following the same naming convention with~\cite{li2020gfte,chi2019complicated,qasim2019rethinking}, the connecting relations can be divided into horizontal and vertical types.  The main idea is that if two aligned bounding boxes has enough overlap in x/y-coordinate, we will match them in vertical/horizontal direction.  Mathematically, for every two aligned bounding boxes, $\{(x_{1}, y_{1}),(x_{2}, y_{2})\}$ and $\{(x'_{1}, y'_{1}),(x'_{2}, y'_{2})\}$, they will be horizontally connected if $y'_{1} \le \frac{y_{1}+y_{2}}{2}$$\le$$y'_{2}$ or $y_{1}$$\le$$\frac{y'_{1}+y'_{2}}{2}$$\le$$y_{2}$. Similarly, they will be vertically connected if  $x'_{1}$$\le$$\frac{x_{1}+x_{2}}{2}$$\le x'_{2}$ or $x_{1}$$\le$$\frac{x'_{1}+x'_{2}}{2}$$\le$$x_{2}$.

\emph{Empty Cell Searching. }
After obtaining the relations between the detected aligned bounding boxes, we treat them as nodes in a graph, and the connected relations are edges. All of the nodes in the same row/column make up a complete subgraph. Inspired by \cite{qasim2019rethinking},
we adopt the algorithm of Maximum Clique Search \cite{bron1973algorithm} to find all maximum cliques in the graph. Take the row searching process as an example, every node that belongs to the same row will be in the same clique.
For the cell that crosses span multiple rows, the corresponding node will appear multiple times in different cliques.
After sorting these cliques by the average $y$-coordinate, we can easily label each node with its row index. Nodes that appear in multiple cliques will be labeled with multiple row indices. We can easily find those vacant positions, which are corresponding to the empty cells.

\emph{Empty Cell Merging.}
By now, we have obtained the empty cells at the smallest level (occupies 1 row and 1 column).
To merge these cells more feasibly, we first assign the single empty cells with the aligned bounding box shape as the cell's maximum height/width in the same row/column.  Thanks to the visual clues learned by the global segmentation task, we can design the simple merging strategy following the segmentation result. We compute the ratio of pixels that are predicted as 1 in the interval region for every two neighbor empty cells, as the red region illustrated in Figure \ref{fig:4}. If the ratio is larger than the preset threshold, we will merge these two cells.
As we can see, the empty regions' visual ambiguity always exists, and the segmentation task can hardly be learned perfectly. That is why many segmentation-based methods~\cite{qasim2019rethinking,prasad2020cascadetabnet,paliwal2019tablenet} struggle with complicated post-processing, such as fracture completion and threshold setting. The proposed method straightforwardly adopts the original visual clue provided by global segmentation and uses pixel voting to obtain a more reliable result.

\section{Experiments}

\subsection{Datasets}
We evaluate our proposed framework on following popular benchmarks that contain the annotations of both text content bounding boxes and cell relations.

\textbf{ICDAR 2013}~\cite{gobel2013icdar}. This dataset contains 98 training samples and 156 testing samples that cropped from the PDF of government reports.

\textbf{SciTSR}~\cite{chi2019complicated}. This dataset contains 12,000 training images and 3,000 testing images cropped from PDF of scientific literature. Authors also select a subset of complicated samples that contains 2,885 training images and 716 testing images, called SciTSR-COMP.

\textbf{PubTabNet}~\cite{zhong2019image}. It is a large-scale complicated table collection that contains 500,777 training images, 9,115 validating images and 9,138 testing images. This dataset contains a large amount of three-lines tables with multi-row/column cells, empty cells, etc.
\subsection{Implementation Details}
All experiments are implemented in Pytorch with 8$\times$32 GB-Tesla-V100 GPUs.  The deep features are extracted and aggregated through the backbone of ResNet-50~\cite{he2016deep} with Feature Pyramid Network (FPN)~\cite{lin2017feature}. The weights of the backbone are initialized from the pre-trained model of MS-COCO~\cite{LinMBHPRDZ14}. In LPMA, the model generates anchors in six different ratios $[1/20, 1/10, 1/5, 1/2, 1, 2]$ for capturing the different shapes of bounding boxes. The Non-Maximal suppression (NMS) IoU threshold of RCNN is 0.1 in the testing phase.

For all benchmarks, the model is trained by the SGD optimizer with batch-size=4, momentum=0.9, and weight-decay=$1\times 10^{-4}$. The initial learning ratio of $1\times 10^{-2}$ is divided by 10 every 5 epochs.  The model's training on SciTSR and PubTabNet lasts for 12 epochs, and the fine-tuning process on ICDAR 2013 lasts for 25 epochs. We also randomly scale the longer side of the input images to the lengths in the range [480, 1080] for all training processes. In the testing phase, we set the longer side of the input image as 768.
We empirically set all weight parameters as $\lambda_1$$=$$\lambda_2$$=$$\lambda_3$$=$$1$.

\begin{table}[t]
\caption{Results on ICDAR 2013, SciTSR, SciTSR-COMP datasets. P, R, F1 represent Precision, Recall, F1-Score, respectively.
Symbol of $\dagger$ means pre-trained data are used.}
\begin{center}
\resizebox{0.99\textwidth}{!}{
\begin{tabular}{|l|c|c c c| c c c| c c c|}
\hline
\multirow{2}{*}{Methods}  & \multirow{2}{*}{\makecell[c]{Training\\ Dataset}}  & \multicolumn{3}{c|} {ICDAR 2013 } &  \multicolumn{3}{c|}{SciTSR} &  \multicolumn{3}{c|}{SciTSR-COMP }  \\ \cline{3-11}
& & P & R   &  F1  &  P &  R  &  F1  & P&  R &  F1 \\ \cline{4-11}
\hline
DeepDeSRT~\cite{schreiber2017deepdesrt} & - & 0.959 & 0.874 & 0.914 &  0.906 & 0.887 & 0.890 & 0.863 & 0.831 & 0.846\\
Split~\cite{tensmeyer2019deep} & Private  & 0.869 & 0.866 & 0.868 &  - & - & -& - & - & -\\
DeepTabStR~\cite{siddiqui2019deeptabstr}& ICDAR 2013 &0.931 & 0.930 & 0.930 &  - & - & -& - & - & -\\
Siddiqui et al.~\cite{siddiqui2019rethinking} & Synthetic 500k & 0.934 & 0.934 & 0.934 &  - & - & -& - & - & - \\
ReS2TIM~\cite{xue2019res2tim} & ICDAR 2013$\dagger$ & 0.734 & 0.747 & 0.740 &  - & - & -& - & - & -\\
GTE~\cite{zheng2020global}& ICDAR 2013$\dagger$ & 0.944 & 0.927 & 0.935 &  - & - & - & -& - & - \\
GraphTSR~\cite{chi2019complicated} & SciTSR & 0.885 & 0.860 & 0.872 & 0.959 & 0.948 & 0.953 & 0.964 & 0.945 & 0.955\\
TabStruct-Net~\cite{raja2020table} & SciTSR & 0.915 & 0.897 & 0.906 &  0.927 & 0.913 & 0.920 & 0.909 & 0.882 & 0.895  \\
\hline
LGPMA & SciTSR &0.930 & 0.977 & 0.953 & \textbf{0.982} & \textbf{0.993} & \textbf{0.988} & \textbf{0.973} & \textbf{0.987} & \textbf{0.980}  \\
LGPMA & ICDAR 2013$\dagger$ &\textbf{0.967} & \textbf{0.991} & \textbf{0.979} & - & - & - & - & - & -  \\
\hline
\end{tabular}
}
\end{center}
\label{tb1}
\end{table}
\subsection{Results on Table Structure Recognition Benchmarks}
We first conduct experiments on the datasets of ICDAR 2013 and SciTSR, and the evaluation metric follows \cite{gobel2013icdar} (counting the micro-averaged correctness of neighboring relations).
Notice that since ICDAR 2013 has very few samples, many previous works used different training or pre-training data. To be comparable to~\cite{chi2019complicated} and~\cite{raja2020table}, our model is only trained by the training set of SciTSR. We also report the result of the model that is then fine-tuned on the training set of ICDAR2013, labeled by $\dagger$. The results of DeepDeSRT on SciTSR come from~\cite{chi2019complicated}.
The results are demonstrated in Table \ref{tb1}, from which we can see that the proposed LGP-TabNet vastly surpasses previous advances on these three benchmarks by 4.4\%, 3.5\%, 2.5\%, respectively. Beside, LGPMA shows no performance decline on the complicated testing dataset SciTSR-COMP, which demonstrates its powerful ability to perceive spatial relations.

We also test our model in a more challenging benchmark of PubTabNet, whose results are demonstrated in Table \ref{tb2}. Since the corresponding evaluation metric of TEDS~\cite{zhong2019image} considers both table structure and text content, we simply adopt an attention-based model~\cite{DBLP:conf/cvpr/LeeO16} to recognize the text recognition. In the results, our method surpasses the previous SOTA by 1.6 in TEDS.  We also report the results that only considers the table structure, denoted as TEDS-Struc. The performance gap between TEDS-Struc and TEDS mainly comes from the recognition error and annotation ambiguities.

\begin{table}[t]
\caption{Results on PubTabNet. TESDS-Struc only considers the table structures.}
\begin{center}
\scalebox{0.95}{
\begin{tabular}{|l|c|c|c|c|}
\hline
{Methods} &  \makecell[c]{Training \\Dataset} &  \makecell[c]{Tesing \\Dataset}  & \makecell[c]{TEDS \\(All)} & \makecell[c]{TEDS-Struc. \\(All)} \\ \cline{1-5}
EDD~\cite{zhong2019image} & PTN-train & PTN-val &  88.3 & - \\
TabStruct-Net~\cite{raja2020table} & SciTSR & PTN-val & 90.1 & - \\
GTE~\cite{zheng2020global} &PTN-train & PTN-val & 93.0 & -  \\
\hline
LGPMA (ours) & PTN-train & PTN-val &  \textbf{94.6} & \textbf{96.7} \\
\hline
\end{tabular}
}
\end{center}
\label{tb2}
\end{table}
\subsubsection{Visualization Results.}
\begin{figure*}[t]
\begin{center}
\includegraphics[width=1\textwidth,]{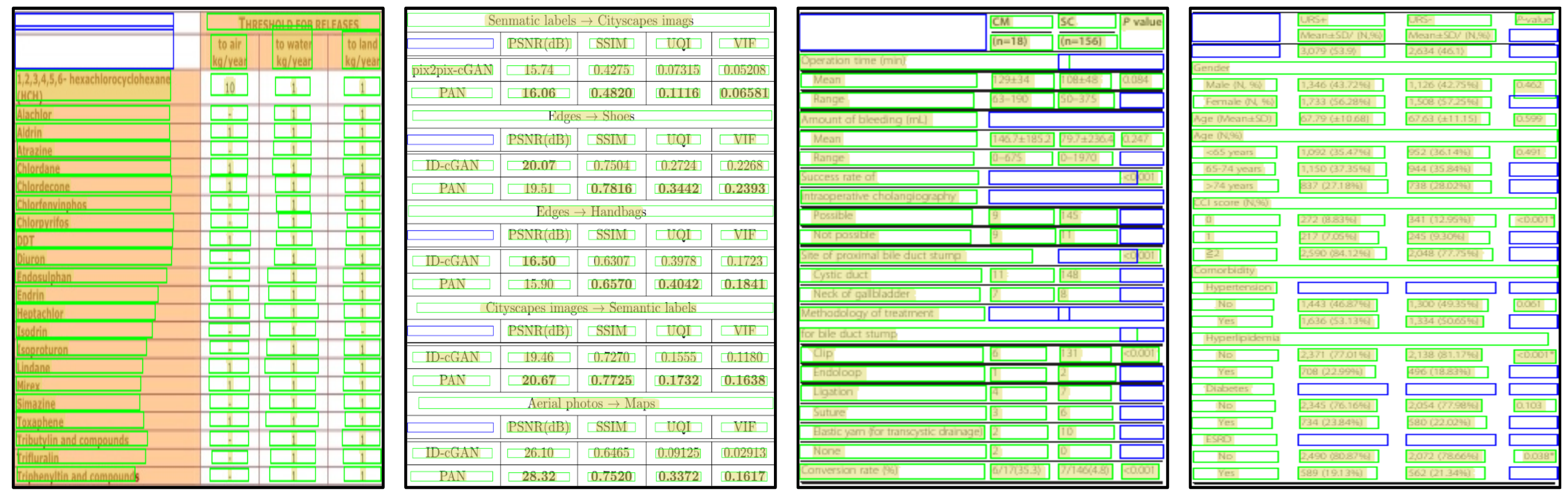}\\
\end{center}
\caption{Visualization results on ICDAR2013, SciTSR, PubTabNet. Green boxes are detected aligned bounding boxes, and blue boxes are empty cells generated by the proposed table structure recovery pipeline.}
\label{vis1}
\end{figure*}

We demonstrate some of the visualization results in Figure \ref{vis1}, in which the green boxes denote the predicted non-empty boxes, and blue boxes are the recovered empty boxes. We can see that our model can predict cells' accurate boundaries, even for those cross span multiple rows/columns.

\begin{figure*}[t]
\begin{center}
\includegraphics[width=1\textwidth,]{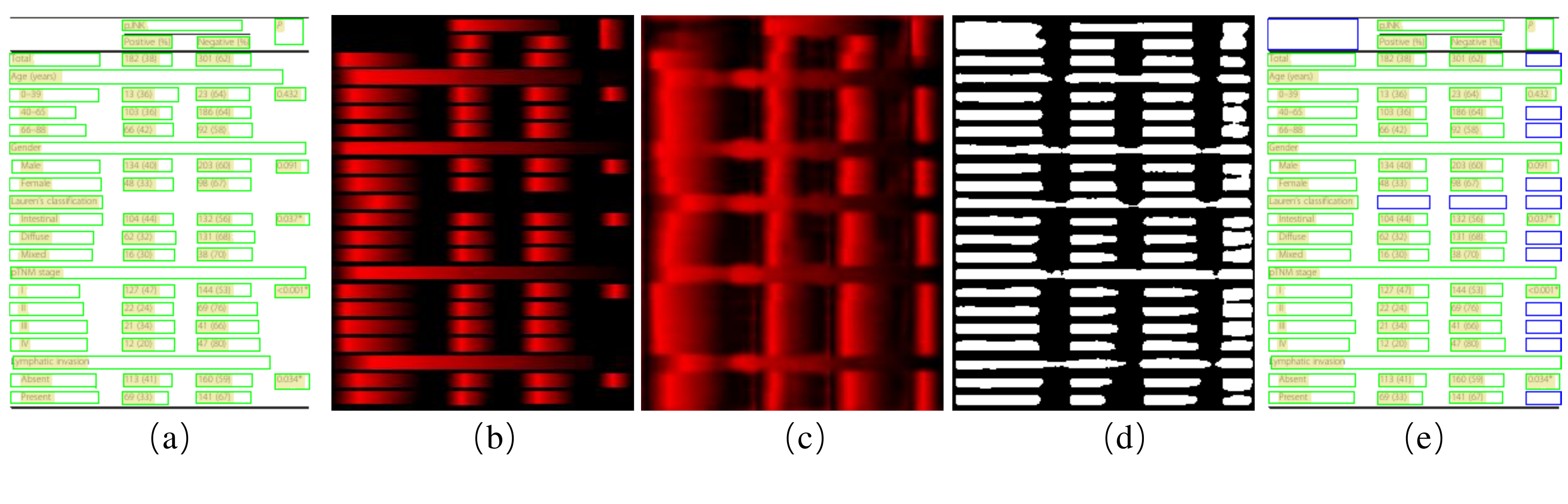}\\
\end{center}
\caption{Visualization of an example that is successfully refined.  (a) The aligned bounding boxes before refinement. (b) LPMA (in horizontal). (c) GPMA (in horizontal). (d) Global binary segmentation. (e) Final result after refinement and empty cell merging.}
\label{vis2}
\end{figure*}

Figure \ref{vis2} demonstrates an example that is successfully predicted due to the correct refinement. We only show the LPMA and GPMA maps in the horizontal direction, where the LPMA map is generated by overlapping all proposals' local maps. In the initially proposed bounding boxes, some boxes do not have enough breadth, which would lead to wrong matching results. After refining by LPMA and GPMA, these boundaries can reach the more feasible positions. The empty cells can also be merged feasibly according to the predicted segmentation map during the table structure recovery.

\subsection{Ablation Studies}
We randomly select 60,000 training images and 1,000 validation images from PubTabNet to conduct the subsequent ablation studies.
\subsubsection{Effectiveness of Aligned Bounding Box Detection.}
To verify the effectiveness of the designed modules, we conduct a group of ablation experiments, as shown in Table \ref{tb3}. Besides the TEDS-Struc metric, we also report the text region's detection results and aligned bounding boxes, where the detection IoU threshold is 0.7, and empty cells are ignored. From the results, we can easily find that both LPMA and GPMA can vastly enhance the aligned bounding box detection performance, which is also proportional to the performance of TEDS-Struc. Although these modules are designed for aligned bounding boxes, they also slightly enhance the performance of text region detection results.
The performance gap between text region detection and aligned bounding box detection again demonstrates the latter task is much more difficult and challenging.

We also evaluate the effectiveness of Alignment Loss (abbr. AL)  proposed by~\cite{raja2020table}. Although solely adopting AL achieves better performance than the original Mask-RCNN, the performance is even lower than the best results of LGPMA when compromising all three modules. It means AL might bring adverse impact on LGPMA. Compared to AL, our proposed LGPMA can obtain more performance gain by 3.1\% in aligned bounding box detection and 0.59 in TEDS-Struc.

\begin{table}[t]
\caption{Ablation experiments on that how different modules effect the aligned bounding box detection}
\begin{center}
\resizebox{0.99\textwidth}{!}{
\begin{tabular}{|l|c c c|c c c | c c c|c| }
\hline
\multirow{2}{*}{Models} &  \multicolumn{3}{c|}{Modules} &  \multicolumn{3}{c|}{Det of text regions} &   \multicolumn{3}{c|}{ \makecell[c]{Det of non-empty \\aligned bounding boxes}} & \multirow{2}{*}{\makecell{TEDS- \\Struc.}}  \\ \cline{2-10}
&LPMA & GPMA & AL\cite{raja2020table} & Precision & Recall & Hmean & Precision & Rrecall & Hmean & \\ \cline{2-11}
\hline
Faster R-CNN & & & & - & - & - & 81.32 & 81.31 & 81.31 & 94.63 \\
\hline
\multirow{4}{*}{Mask R-CNN} & & & & 91.71 & 91.53 & 91.62 & 81.83 & 81.82 & 81.83 & 94.65 \\
 & \ding{51} &  & & 91.92 & 91.66 & 91.79 & 84.29 & 84.10 & 84.20 & 95.22 \\
 & & \ding{51} & & 91.98 & 91.50 & 91.74 & 83.48 & 83.18 & 83.33 & 95.04 \\
 &\ding{51} & \ding{51} & & \textbf{92.27} & \textbf{91.86} & \textbf{92.06} & \textbf{85.14} &\textbf{84.77} & \textbf{84.95} & \textbf{95.53} \\
\hline
\multirow{2}{*}{Mask R-CNN} & &  & \ding{51} & 92.11 & 91.85 & 91.98 & 81.91 & 81.79 & 81.85 & 94.94 \\
 &\ding{51} & \ding{51} & \ding{51} & 92.05 & 91.65 & 91.85 & 84.87 & 84.50 & 84.68 & 95.31 \\
\hline
\end{tabular}
}
\end{center}
\label{tb3}
\end{table}
\begin{table}[t]
\caption{Ablation experiments on different empty cells merging strategy. The bottom part shows the results given non-empty aligned bounding boxes ground-truth.}
\begin{center}
\resizebox{0.99\textwidth}{!}{
\begin{tabular}{|l|c|c c c | c c c| c| }
\hline
\multirow{2}{*}{\makecell[c]{Structure recovery \\ strategies}} & \multirow{2}{*}{\makecell[c]{with non-empty\\ box GT?}} &\multicolumn{3}{c|}{\makecell[c]{Det of empty \\aligned bounding boxes}} &   \multicolumn{3}{c|}{\makecell[c]{Det of all \\ aligned bounding boxes}} & \multirow{2}{*}{\makecell{TEDS- \\Struc.}}   \\ \cline{3-8}
& & Precision & Recall & Hmean & Precision & Recall & Hmean & \\ \cline{3-8}
\hline
Minimum empty cells &\ding{55} & 56.17 & 63.04 & 59.40 & 82.36 & 82.74 & 82.55 & 95.50 \\
Maximum empty cells & \ding{55}& 23.19 & 10.76 & 14.70 & 82.30 & 77.39 & 79.77 & 92.57 \\
Proposed LGPMA & \ding{55} & \textbf{68.12} & \textbf{72.43} & \textbf{70.21} & \textbf{82.69} & \textbf{83.17} & \textbf{82.93} & \textbf{95.53}\\
\hline
Minimum cells & \ding{51}& 96.14 & 97.00 & 96.56 & 99.60 & 99.69 & 99.64 & 99.52 \\
Maximum cells & \ding{51}& 41.13 & 15.22 & 22.22 & 97.58 & 91.19 & 94.28 & 95.60 \\
Proposed LGPMA &\ding{51} & \textbf{97.26} & \textbf{97.68} & \textbf{97.47} & \textbf{99.85} & \textbf{99.88} & \textbf{99.87} & \textbf{99.77} \\
\hline
\end{tabular}
}
\end{center}
\label{tb4}
\end{table}

\subsubsection{Effectiveness of Table Structure Recovery.}
To verify the proposed table recovery pipeline's effectiveness, we conduct experiments to compare different empty cell merging strategies, as illustrated in Table \ref{tb4}. The evaluations contain the detection results that only consider empty cells, the detection results of all aligned bounding boxes, and TEDS-Struc. The strategy of using minimum cells means the direct results after \emph{Empty Cell Searching}, and using maximum cells means merging all neighboring empty cells with the same height/width. From the results, we can see that our strategy using the visual information from GPMA can correctly merge many empty cells and obtain the highest performances in both detection and TED-Struc metrics. Compared with Strategy of \emph{Minimum empty cells }, the promotion of \emph{Empty Cell Merging by LGPMA} on TEDS-Struc is relatively small. This is because the number of empty cells minority, and most of them in this dataset are labeled in the smallest shapes. Nevertheless, our proposed empty cell merging strategy is more robust to adapt to any possibility.

Suppose the aligned bounding boxes of non-empty cells are detected perfectly, which equals the situation given ground-truths, as shown in the bottom part of Table \ref{tb4}. In this case, we can easily find that using the strategy of whether \emph{Minimum cells} or the proposed \emph{LGPMA} can almost achieve 100\% accuracy in table structure recovery, and many errors come from the noisy labels. It demonstrates the robustness of our table structure recovery pipeline, and the performance mainly depends on the correctness of aligned bounding box detection.

\section{Conclusion}
\label{conclusion}
In this paper, we present a novel framework for table structure recognition named LGPMA. We adopt the local and global pyramid mask learning to compromises advantages from both local texture and global layout information. In the inference stage, fusing the two levels' predictions via a mask re-scoring strategy, the network generates more reliable aligned bounding boxes.
Finally, we propose a uniform table structure recovery pipeline to get the final results, which can also predict the feasible empty cell partition. Experimental results demonstrate our method has achieved the new state-of-the-art in three public benchmarks.

\bibliographystyle{splncs04}
\bibliography{ref_mini}

\begin{thebibliography}{10}
\providecommand{\url}[1]{\texttt{#1}}
\providecommand{\urlprefix}{URL }
\providecommand{\doi}[1]{https://doi.org/#1}

\bibitem{bron1973algorithm}
Bron, C., Kerbosch, J.: Finding all cliques of an undirected graph (algorithm
  457). Commun. {ACM}  \textbf{16}(9),  575--576 (1973)

\bibitem{chi2019complicated}
Chi, Z., Huang, H., Xu, H., Yu, H., Yin, W., Mao, X.: Complicated table
  structure recognition. CoRR  \textbf{abs/1908.04729} (2019)

\bibitem{DoushP10}
Doush, I.A., Pontelli, E.: Detecting and recognizing tables in spreadsheets.
  In: IAPR. pp. 471--478 (2010)

\bibitem{gao2019icdar}
Gao, L., Huang, Y., D{\'{e}}jean, H., Meunier, J., Yan, Q., Fang, Y., Kleber,
  F., Lang, E.M.: {ICDAR} 2019 competition on table detection and recognition
  (ctdar). In: ICDAR. pp. 1510--1515 (2019)

\bibitem{gobel2013icdar}
G{\"{o}}bel, M.C., Hassan, T., Oro, E., Orsi, G.: {ICDAR} 2013 table
  competition. In: ICDAR. pp. 1449--1453 (2013)

\bibitem{he2017mask}
He, K., Gkioxari, G., Doll{\'{a}}r, P., Girshick, R.B.: Mask {R-CNN}. In: ICCV.
  pp. 2980--2988 (2017)

\bibitem{he2016deep}
He, K., Zhang, X., Ren, S., Sun, J.: Deep residual learning for image
  recognition. In: CVPR. pp. 770--778 (2016)

\bibitem{Itonori93}
Itonori, K.: Table structure recognition based on textblock arrangement and
  ruled line position. In: ICDAR. pp. 765--768 (1993)

\bibitem{khan2019table}
Khan, S.A., Khalid, S.M.D., Shahzad, M.A., Shafait, F.: Table structure
  extraction with bi-directional gated recurrent unit networks. In: ICDAR. pp.
  1366--1371 (2019)

\bibitem{kieninger1998table}
Kieninger, T.: Table structure recognition based on robust block segmentation.
  In: Document Recognition V. vol.~3305, pp. 22--32 (1998)

\bibitem{KociTL018}
Koci, E., Thiele, M., Lehner, W., Romero, O.: Table recognition in spreadsheets
  via a graph representation. In: IAPR. pp. 139--144 (2018)

\bibitem{DBLP:conf/cvpr/LeeO16}
Lee, C., Osindero, S.: Recursive recurrent nets with attention modeling for
  {OCR} in the wild. In: CVPR. pp. 2231--2239 (2016)

\bibitem{LiCHWZL20}
Li, M., Cui, L., Huang, S., Wei, F., Zhou, M., Li, Z.: Tablebank: Table
  benchmark for image-based table detection and recognition. In: LREC. pp.
  1918--1925 (2020)

\bibitem{li2020gfte}
Li, Y., Huang, Z., Yan, J., Zhou, Y., Ye, F., Liu, X.: {GFTE:} graph-based
  financial table extraction. In: ICPR Workshops. vol. 12662, pp. 644--658
  (2020)

\bibitem{lin2017feature}
Lin, T., Doll{\'{a}}r, P., Girshick, R.B., He, K., Hariharan, B., Belongie,
  S.J.: Feature pyramid networks for object detection. In: CVPR. pp. 936--944
  (2017)

\bibitem{LinMBHPRDZ14}
Lin, T., Maire, M., Belongie, S.J., Hays, J., Perona, P., Ramanan, D.,
  Doll{\'{a}}r, P., Zitnick, C.L.: Microsoft {COCO:} common objects in context.
  In: ECCV. Lecture Notes in Computer Science, vol.~8693, pp. 740--755 (2014)

\bibitem{liu2019pyramid}
Liu, J., Liu, X., Sheng, J., Liang, D., Li, X., Liu, Q.: Pyramid mask text
  detector. CoRR  \textbf{abs/1903.11800} (2019)

\bibitem{liu2009improving}
Liu, Y., Bai, K., Mitra, P., Giles, C.L.: Improving the table boundary
  detection in pdfs by fixing the sequence error of the sparse lines. In:
  ICDAR. pp. 1006--1010 (2009)

\bibitem{liu2008identifying}
Liu, Y., Mitra, P., Giles, C.L.: Identifying table boundaries in digital
  documents via sparse line detection. In: CIKM. pp. 1311--1320 (2008)

\bibitem{milletari2016v}
Milletari, F., Navab, N., Ahmadi, S.: V-net: Fully convolutional neural
  networks for volumetric medical image segmentation. In: 3DV. pp. 565--571
  (2016)

\bibitem{nishida2017understanding}
Nishida, K., Sadamitsu, K., Higashinaka, R., Matsuo, Y.: Understanding the
  semantic structures of tables with a hybrid deep neural network architecture.
  In: AAAI. pp. 168--174 (2017)

\bibitem{paliwal2019tablenet}
Paliwal, S.S., D, V., Rahul, R., Sharma, M., Vig, L.: Tablenet: Deep learning
  model for end-to-end table detection and tabular data extraction from scanned
  document images. In: ICDAR. pp. 128--133 (2019)

\bibitem{prasad2020cascadetabnet}
Prasad, D., Gadpal, A., Kapadni, K., Visave, M., Sultanpure, K.: Cascadetabnet:
  An approach for end to end table detection and structure recognition from
  image-based documents. In: CVPR Workshops. pp. 2439--2447 (2020)

\bibitem{qasim2019rethinking}
Qasim, S.R., Mahmood, H., Shafait, F.: Rethinking table recognition using graph
  neural networks. In: ICDAR. pp. 142--147 (2019)

\bibitem{qiao2020text}
Qiao, L., Tang, S., Cheng, Z., Xu, Y., Niu, Y., Pu, S., Wu, F.: Text
  perceptron: Towards end-to-end arbitrary-shaped text spotting. In: AAAI. pp.
  11899--11907 (2020)

\bibitem{raja2020table}
Raja, S., Mondal, A., Jawahar, C.V.: Table structure recognition using top-down
  and bottom-up cues. In: ECCV. Lecture Notes in Computer Science, vol. 12373,
  pp. 70--86 (2020)

\bibitem{Redmon2016You}
Redmon, J., Divvala, S.K., Girshick, R.B., Farhadi, A.: You only look once:
  Unified, real-time object detection. In: CVPR. pp. 779--788 (2016)

\bibitem{2015Faster}
Ren, S., He, K., Girshick, R.B., Sun, J.: Faster {R-CNN:} towards real-time
  object detection with region proposal networks. In: NeurIPS. pp. 91--99
  (2015)

\bibitem{ScarselliGTHM09}
Scarselli, F., Gori, M., Tsoi, A.C., Hagenbuchner, M., Monfardini, G.: The
  graph neural network model. {IEEE} Trans. Neural Networks  \textbf{20}(1),
  61--80 (2009)

\bibitem{schreiber2017deepdesrt}
Schreiber, S., Agne, S., Wolf, I., Dengel, A., Ahmed, S.: Deepdesrt: Deep
  learning for detection and structure recognition of tables in document
  images. In: ICDAR. pp. 1162--1167 (2017)

\bibitem{siddiqui2019deeptabstr}
Siddiqui, S.A., Fateh, I.A., Rizvi, S.T.R., Dengel, A., Ahmed, S.: Deeptabstr:
  Deep learning based table structure recognition. In: ICDAR. pp. 1403--1409
  (2019)

\bibitem{siddiqui2019rethinking}
Siddiqui, S.A., Khan, P.I., Dengel, A., Ahmed, S.: Rethinking semantic
  segmentation for table structure recognition in documents. In: ICDAR. pp.
  1397--1402 (2019)

\bibitem{tensmeyer2019deep}
Tensmeyer, C., Morariu, V.I., Price, B.L., Cohen, S., Martinez, T.R.: Deep
  splitting and merging for table structure decomposition. In: ICDAR. pp.
  114--121 (2019)

\bibitem{WangPH04}
Wang, Y., Phillips, I.T., Haralick, R.M.: Table structure understanding and its
  performance evaluation. Pattern Recognit.  \textbf{37}(7),  1479--1497 (2004)

\bibitem{xie2018scene}
Xie, E., Zang, Y., Shao, S., Yu, G., Yao, C., Li, G.: Scene text detection with
  supervised pyramid context network. In: AAAI. pp. 9038--9045 (2019)

\bibitem{xue2019res2tim}
Xue, W., Li, Q., Tao, D.: Res2tim: Reconstruct syntactic structures from table
  images. In: ICDAR. pp. 749--755 (2019)

\bibitem{ZanibbiBC04}
Zanibbi, R., Blostein, D., Cordy, J.R.: A survey of table recognition. Int. J.
  Document Anal. Recognit.  \textbf{7}(1),  1--16 (2004)

\bibitem{zheng2020global}
Zheng, X., Burdick, D., Popa, L., Wang, N.X.R.: Global table extractor {(GTE):}
  {A} framework for joint table identification and cell structure recognition
  using visual context. CoRR  \textbf{abs/2005.00589} (2020)

\bibitem{zhong2019image}
Zhong, X., ShafieiBavani, E., Jimeno{-}Yepes, A.: Image-based table
  recognition: Data, model, and evaluation. In: ECCV. vol. 12366, pp. 564--580
  (2020)

\bibitem{zhou2017east}
Zhou, X., Yao, C., Wen, H., Wang, Y., Zhou, S., He, W., Liang, J.: {EAST:} an
  efficient and accurate scene text detector. In: CVPR. pp. 2642--2651 (2017)

\end{thebibliography}

\end{document}